\documentclass[10pt,twocolumn,letterpaper]{article}

\usepackage[pagenumbers]{cvpr} % To force page numbers, e.g. for an arXiv version

\usepackage{amsmath,amssymb,amsfonts}
\usepackage{graphicx}
\usepackage{booktabs}
\usepackage{multirow}
\usepackage{xcolor}
\usepackage{subcaption}
\usepackage{tabularx}
\usepackage{tikz}
\usetikzlibrary{shapes.geometric,arrows.meta,positioning,fit,backgrounds,calc,decorations.pathreplacing}

\newcommand{\xzero}{\mathbf{x}_0}
\newcommand{\xt}{\mathbf{x}_t}
\newcommand{\cc}{\mathbf{c}}

\definecolor{cvprblue}{rgb}{0.21,0.49,0.74}
\usepackage[pagebackref,breaklinks,colorlinks,allcolors=cvprblue]{hyperref}

\def\paperID{*****} % *** Enter the Paper ID here
\def\confName{CVPR}
\def\confYear{2026}

\title{Toward Phonology-Guided Sign Language Motion Generation:\\A Diffusion Baseline and Conditioning Analysis}

\author{Rui Hong, Jana Ko\v{s}eck\'{a} \\
George Mason University \\
{\tt\small \{rhong5, kosecka\}@gmu.edu}
}

\begin{document}
\maketitle

% ======================================================================
% ABSTRACT
% ======================================================================
\begin{abstract}

Generating natural, correct, and visually smooth 3D avatar sign language motion conditioned on the text inputs continues to be very challenging. 
In this work, we train a generative model of 3D body motion and explore the role
of phonological attribute conditioning for sign language motion generation, using ASL-LEX 2.0 annotations such as hand shape, hand location and movement. We first establish a strong diffusion baseline using an Human Motion MDM-style~\cite{tevet2023human} diffusion model with SMPL-X representation, which outperforms SignAvatar~\cite{dong2024signavatar}, a state-of-the-art CVAE method, on gloss discriminability metrics. We then systematically study the role of text conditioning
using different text encoders (CLIP vs.\ T5), conditioning modes (gloss-only vs.\ gloss+phonological attributes), and attribute notation format (symbolic vs.\ natural language). 
Our analysis reveals that translating symbolic ASL-LEX notations to natural language is a necessary condition for effective CLIP-based attribute conditioning, while T5 is largely unaffected by this translation. 
Furthermore, our best-performing variant (CLIP with mapped attributes) outperforms SignAvatar across all metrics.
These findings highlight input representation as a critical factor for text-encoder-based attribute conditioning, and motivate structured conditioning approaches where gloss and phonological attributes are encoded through independent pathways.
Code is available at \url{https://github.com/hongrui16/phonology-guided-sign-gen}.

% {\bf JK discuss} 
% We hypothesize two contributing factors for the conditioning failure: noisy alignment between WLASL video variants and ASL-LEX canonical forms, and shared attributes collapsing gloss identity. 
%These findings provide concrete motivation for structured conditioning, where gloss and phonological attributes are encoded through independent pathways, which we outline as a promising direction for compositional sign language generation.
\end{abstract}

% ======================================================================
% 1. INTRODUCTION
% ======================================================================
\section{Introduction}
\label{sec:intro}

Sign language is the primary mode of communication for millions of deaf or  hard-of-hearing individuals worldwide. Automatic generation of natural-looking and correct sign language motion has broad applications in accessible communication, virtual signing avatars, and sign language education. This, however, remains a challenging problem due to the complex movement of hand shapes at different hand locations, palm orientation, as well as co-articulated face expressions. 

Recent work on sign language generation has made progress using generative models conditioned on gloss labels~\cite{dong2024signavatar}. However, these approaches typically 
% treat each sign as a black-box token 
% an opaque label, {\bf JK ???} 
ignore the internal phonological structure that linguists have long recognized as fundamental to sign language organization~\cite{stokoe1960sign,brentari1998prosodic}. Just as spoken language phonology decomposes words into phonemes, sign language phonology decomposes signs into a small set of articulatory parameters: hand shape, location, movement, palm orientation, and non-manual signals. The ASL-LEX 2.0 database~\cite{sehyr2021asl} provides these structured annotations for over 2,700 ASL signs across 22 phonological dimensions.

Incorporating phonological structure into sign generation offers compelling advantages. First, it enables \textit{compositional generalization}: a model that understands individual phonological dimensions can potentially generate novel sign-like motions by recombining known attributes. Second, it provides \textit{interpretable control}: users or downstream systems could independently adjust handshape, location, or movement parameters. Third, it introduces a \textit{linguistically grounded inductive bias} that could improve data efficiency by sharing information across signs with similar phonological profiles.

In this paper, we investigate phonological attribute conditioning for 3D sign language motion generation within a diffusion-based framework. Our contributions are:

\begin{enumerate}
    % \item We establish a strong \textbf{diffusion baseline} for gloss-conditioned sign motion generation using an MDM-style~\cite{tevet2023human} Transformer Encoder denoiser operating on SMPL-X~\cite{pavlakos2019expressive} upper-body motion in 6D rotation representation~\cite{zhou2019continuity}. This baseline outperforms SignAvatar~\cite{dong2024signavatar}, a state-of-the-art CVAE method, on gloss discriminability while achieving comparable distributional quality.
    \item We establish a strong \textbf{diffusion baseline} for gloss-conditioned sign motion generation using an MDM-style~\cite{tevet2023human} Transformer Encoder denoiser operating on SMPL-X~\cite{pavlakos2019expressive} upper-body motion in 6D rotation representation~\cite{zhou2019continuity}. This baseline outperforms SignAvatar~\cite{dong2024signavatar}, a state-of-the-art CVAE method, on gloss discriminability while achieving comparable distributional quality; with mapped phonological attribute conditioning, our model further surpasses SignAvatar across all reported metrics.
    \item We conduct a \textbf{systematic study of phonological attribute
    conditioning} for sign motion generation via an ablation over text
    encoders (CLIP~\cite{radford2021learning} vs.\ T5~\cite{raffel2020exploring}),
    conditioning modes (gloss-only vs.\ gloss+phonological attributes from
    ASL-LEX 2.0), and attribute notation format (symbolic vs.\ natural
    language). We demonstrate that translating symbolic ASL-LEX notations
    to natural language is a necessary condition for effective CLIP-based
    attribute conditioning, and that the two encoders respond asymmetrically
    to this translation. 
    % \item We \textbf{hypothesize two contributing factors} for the conditioning failure---noisy label alignment and shared-attribute identity collapse---and use these findings to motivate \textbf{structured tensor conditioning} as a principled direction for phonology-guided generation. {\bf JK - I would omit this}
\end{enumerate}
The ablation of the conditioning e
% We \textbf{hypothesize two contributing factors} for the conditioning failure---noisy label alignment and shared-attribute identity collapse---and use
The findings presented here motivate the use of phonological attributes as conditioning signals, suggesting a principled direction for phonology-guided generation.

% ======================================================================
% 2. RELATED WORK
% ======================================================================
\section{Related Work}
\label{sec:related}

\paragraph{Sign Language Generation.}
Sign language generation aims to produce natural signing motion from linguistic input. Early approaches used rule-based systems~\cite{huenerfauth2006generating} that synthesized motion using concatenation of primitives.
More recent data-driven approaches use deep learning methods to train generative models of realistic
avatar motion conditioned on text or gloss with increasing naturalness. SignAvatar~\cite{dong2024signavatar} uses a conditional variational autoencoder (CVAE) conditioned on CLIP embeddings of gloss labels to generate SMPL-X motion parameters.
\citet{baltatzis2024neuralsignactors} train a diffusion model with an anatomically-informed GNN encoder and autoregressive LSTM decoder, conditioned on frozen CLIP sentence embeddings, on a large-scale 3D-annotated version of How2Sign to generate continuous sentence-level ASL motion from free-form text.
Ham2Pose~\cite{arkushin2023ham2pose} generates pose sequences from HamNoSys notation. Progressive Transformer~\cite{saunders2020progressive} and its extensions~\cite{saunders2021signing} address sentence-level sign language production from text. 
However, these methods do not exploit  
% as black-box tokens 
% opaque labels {\bf JK ??} 
internal structure, or instead rely on transcription systems (HamNoSys) that require expert annotation. Our work bridges this gap by conditioning generation on phonological attributes from ASL-LEX 2.0, providing a structured linguistic description for a large gloss vocabulary.

\paragraph{Human Motion Diffusion Models.}
Diffusion models have achieved strong results in human motion generation. Human Motion Diffusion Model MDM~\cite{tevet2023human} uses a Transformer Encoder denoiser with classifier-free guidance for text-conditioned motion generation. MotionDiffuse~\cite{zhang2022motiondiffuse} applies diffusion to body-part-level motion control. Motion Latent-based Diffusion Model MLD~\cite{chen2023executing} introduces a latent diffusion approach for efficient motion generation. 
For hand and sign motion specifically, \citet{bensabath2025text} train a text-conditioned hand motion diffusion model (HandMDM) on large-scale sign language data, where sign categories are automatically translated into natural language motion descriptions via an LLM, demonstrating cross-domain generalization to unseen sign categories and cross-lingual signs.
Our work builds on the MDM architecture but focuses specifically on sign language generation with structured phonological conditioning, and provides a systematic analysis of conditioning mechanisms.

\paragraph{Sign Language Phonology and ASL-LEX.}
Sign language phonology, pioneered by Stokoe~\cite{stokoe1960sign} and refined by Brentari~\cite{brentari1998prosodic}, decomposes signs into simultaneous articulatory parameters. ASL SignBank~\cite{hochgesang2026aslsignbank} is a widely used community resource that documents a large vocabulary of ASL signs with video examples and categorical linguistic annotations. The original ASL-LEX database~\cite{caselli2017asl} introduced a quantitative lexical resource for ASL combining behavioral frequency measures with phonological coding. ASL-LEX 2.0~\cite{sehyr2021asl} substantially expanded this to 2,723 ASL signs annotated across 22 phonological dimensions including handshape, selected fingers, flexion, thumb position, sign type, major location, minor location, path movement, and others. While ASL SignBank covers a larger vocabulary, ASL-LEX 2.0 provides richer phonological attributes per sign, making it more suitable as a conditioning signal for generative models. These annotations have been widely used in psycholinguistic research but remain underexplored in sign language generation. This work represents, to the best of our knowledge, the first systematic study of ASL-LEX phonological attributes as conditioning inputs for 3D motion generation.

% ======================================================================
% 3. METHOD
% ======================================================================
\section{Method}
\label{sec:method}

We present a diffusion-based framework for sign language motion generation conditioned on gloss labels and, optionally, phonological attributes. We train a transformer-based diffusion model on a subset of the WLASL dataset 
with gloss annotations and explore different strategies for text conditioning. 
Figure~\ref{fig:architecture} provides an overview of the proposed model.

\subsection{Motion Representation}
\label{sec:motion_rep}

Similarly to previous works, we represent the articulated human body using SMPL-X~\cite{pavlakos2019expressive} body model parameters. Since sign language primarily involves the upper body, we operate on a subset of 43 upper-body joints (excluding the root, lower body, and jaw), comprising the spine, torso, arms, and both hands. Each joint rotation is parameterized in 6D continuous rotation representation~\cite{zhou2019continuity}, yielding a per-frame feature vector of dimension $D = 43 \times 6 = 258$. Sequences are sampled at a fixed length of $T = 60$ frames. The root joint is excluded and normalized to a canonical orientation in preprocessing, so the model focuses on relative upper-body articulation.

\subsection{Diffusion Framework}
\label{sec:diffusion}

We adopt the denoising diffusion framework following the Human Motion Diffusion Model (MDM)~\cite{tevet2023human}, with $\xzero$-prediction and a cosine noise schedule~\cite{nichol2021improved}.

% \paragraph{Forward process.}
Given a clean motion sequence $\xzero \in \mathbb{R}^{T \times D}$, the forward process adds Gaussian noise over $N = 1000$ time steps:
\begin{equation}
    q(\xt | \xzero) = \mathcal{N}(\xt; \sqrt{\bar{\alpha}_t}\,\xzero, (1 - \bar{\alpha}_t)\,\mathbf{I}),
\end{equation}
where $\bar{\alpha}_t = \prod_{s=1}^{t} (1 - \beta_s)$ and $\{\beta_t\}$ follows the cosine schedule.

%\paragraph{Denoiser architecture.}
The denoiser $f_\theta(\xt, t, \cc)$ predicts $\xzero$ from the noisy input $\xt$, timestep $t$, and condition $\cc$. It consists of a transformer encoder operating over a sequence of tokens:
\begin{equation}
    [\mathbf{h}_c, \mathbf{h}_t, \mathbf{h}_1, \ldots, \mathbf{h}_T] = \text{TransEnc}([\mathbf{e}_c, \mathbf{e}_t, \mathbf{m}_1, \ldots, \mathbf{m}_T])
\end{equation}
where $\mathbf{e}_t$ is the timestep embedding (sinusoidal encoding projected through a 2-layer MLP), $\mathbf{e}_c$ is the condition embedding, and $\mathbf{m}_i = \text{PoseProj}(\xt^{(i)}) + \text{PE}(i)$ are the motion tokens with learned positional encoding. The transformer encoder has 4 layers, 8 attention heads, model dimension 512, and feedforward dimension 2048 with GELU activation. The output motion tokens are projected back to the input dimension via a 2-layer MLP to produce $\hat{\xzero}$.

\paragraph{Training objective.}
We train the model with a combination of reconstruction and velocity losses, both weighted per body part to emphasize hands and arms:
\begin{equation}
    \mathcal{L} = \sum_{g \in \mathcal{G}} w_g \cdot \mathcal{L}_{\text{MSE}}^g + \lambda_{\text{vel}} \sum_{g \in \mathcal{G}} w_g \cdot \mathcal{L}_{\text{vel}}^g,
\end{equation}
where $\mathcal{G} = \{\text{torso}, \text{arms}, \text{lhand}, \text{rhand}\}$ and the weights $w_g$ are set to 5.0 for arms and hands, and 0.5 for torso, reflecting the relative importance of each body part for sign language articulation. The velocity loss $\mathcal{L}_{\text{vel}}^g = \text{MSE}(\Delta\hat{\xzero}^g, \Delta\xzero^g)$ penalizes temporal differences to encourage smooth motion, with $\lambda_{\text{vel}} = 1.0$. Both losses exclude padded frames 
(i.e., frames appended to reach the fixed sequence length of 60).
% {\bf JK ??}

In inference, we use DDIM sampling~\cite{song2021denoising} with 50 steps and $\eta = 0$ (deterministic) for fast generation from Gaussian noise.

\subsection{Conditioning Mechanism}
\label{sec:conditioning}

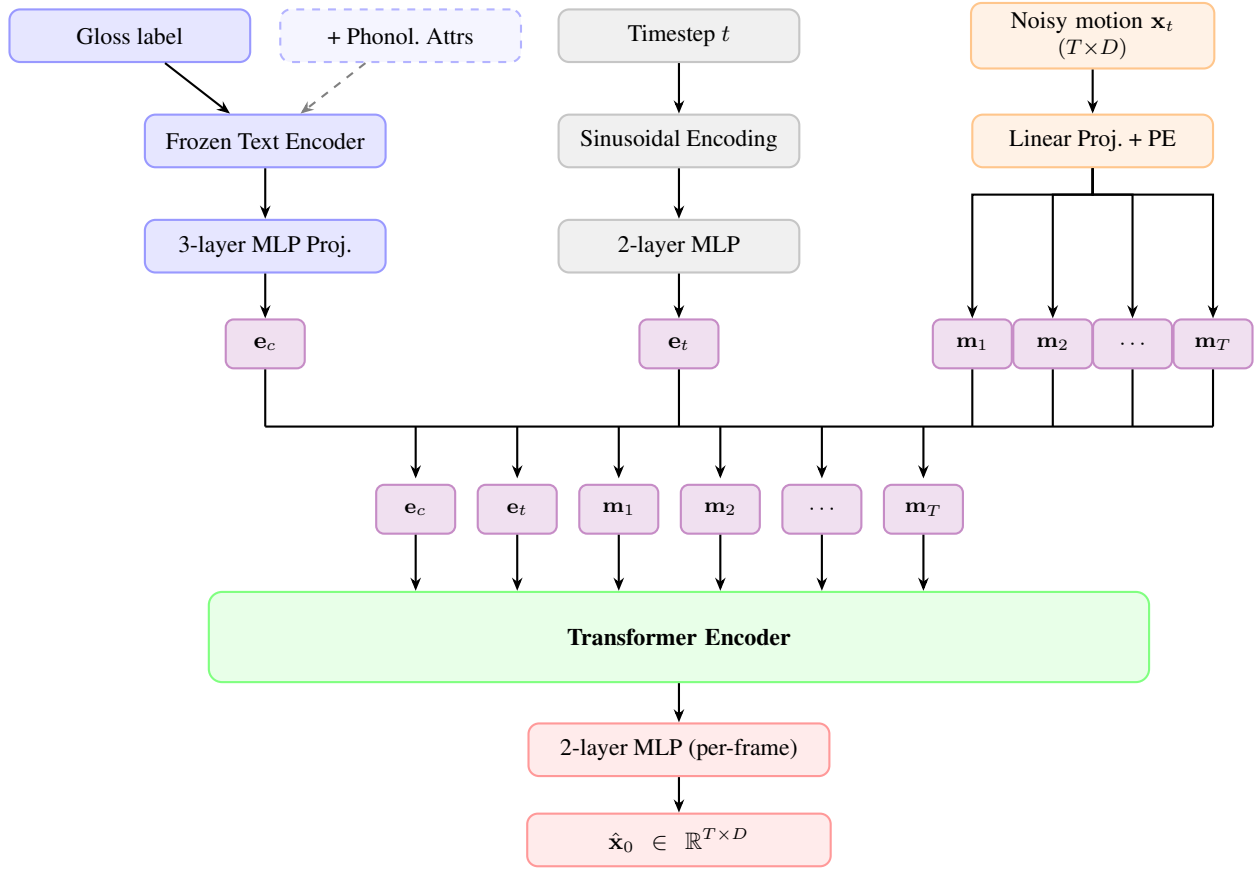
\begin{figure*}[t]
\centering
\begin{tikzpicture}[
    every node/.style={font=\small},
    mainbox/.style={draw, rounded corners=4pt, minimum width=3.2cm, minimum height=0.7cm, text centered, thick},
    condbox/.style={mainbox, fill=blue!10, draw=blue!40},
    timebox/.style={mainbox, fill=gray!12, draw=gray!45},
    motbox/.style={mainbox, fill=orange!10, draw=orange!45},
    tokbox/.style={draw, rounded corners=3pt, fill=violet!12, draw=violet!45,
                   minimum width=1.05cm, minimum height=0.65cm, text centered, thick},
    transbox/.style={draw, rounded corners=5pt, fill=green!9, draw=green!50,
                     minimum height=1.2cm, text centered, thick},
    outbox/.style={mainbox, fill=red!8, draw=red!40},
    arr/.style={-{Stealth[length=5pt,width=4pt]}, thick},
    darr/.style={-{Stealth[length=5pt,width=4pt]}, thick, dashed, gray},
]

% ── Column centres ──────────────────────────────────────────────
\def\cx{0}    % conditioning
\def\tx{5.5}  % timestep
\def\mx{11}   % motion

% ── ROW HEIGHTS ─────────────────────────────────────────────────
\def\ra{0}      % top inputs
\def\rb{-1.4}   % encoder
\def\rc{-2.8}   % projection / MLP
\def\rd{-4.1}   % tokens produced by each branch
\def\re{-6.3}   % merged token sequence (increased gap from \rd=-4.1)
\def\rf{-8.0}   % Transformer Encoder
\def\rg{-9.5}   % output MLP
\def\rh{-10.7}  % x0-hat

% ── COLUMN 1: Conditioning ──────────────────────────────────────
% Gloss label and optional attrs side-by-side at same row
\node[condbox] (gin)  at (\cx-1.8, \ra) {Gloss label};
\node[condbox, dashed, draw=blue!30, fill=blue!4]
    (attrs) at (\cx+1.8, \ra) {+ Phonol.\ Attrs};

\node[condbox] (tenc) at (\cx,\rb)  {Frozen Text Encoder};
\node[condbox] (proj) at (\cx,\rc)  {3-layer MLP Proj.};
\node[tokbox]  (ec)   at (\cx,\rd)  {$\mathbf{e}_c$};

\draw[arr] (gin)   -- (tenc);
\draw[darr] (attrs) -- (tenc);
\draw[arr] (tenc) -- (proj);
\draw[arr] (proj) -- (ec);

% side label

% ═══════════════════════════════════════════════════════════════
% COLUMN 2 — Timestep
% ═══════════════════════════════════════════════════════════════
\node[timebox] (tin)  at (\tx,\ra)  {Timestep $t$};
\node[timebox] (sine) at (\tx,\rb)  {Sinusoidal Encoding};
\node[timebox] (tmlp) at (\tx,\rc)  {2-layer MLP};
\node[tokbox]  (et)   at (\tx,\rd)  {$\mathbf{e}_t$};

\draw[arr] (tin)  -- (sine);
\draw[arr] (sine) -- (tmlp);
\draw[arr] (tmlp) -- (et);

% ═══════════════════════════════════════════════════════════════
% COLUMN 3 — Motion
% ═══════════════════════════════════════════════════════════════
\node[motbox, text width=3cm, align=center] (xt) at (\mx,\ra)
      {Noisy motion $\mathbf{x}_t$\\[-1pt]{\footnotesize$(T{\times}D)$}};
\node[motbox] (pp) at (\mx,\rb)  {Linear Proj.\ + PE};

% four motion tokens spread around column 3 centre
\foreach \i/\lbl/\dx in {1/$\mathbf{m}_1$/-1.6, 2/$\mathbf{m}_2$/-0.53,
                          3/${\cdots}$/0.53,    4/$\mathbf{m}_T$/1.6}{
    \node[tokbox] (mt\i) at (\mx+\dx, \rd) {\footnotesize\lbl};
}

\draw[arr] (xt) -- (pp);
\draw[arr] (pp.south) -- ++(0,-0.35) -| (mt1.north);
\draw[arr] (pp.south) -- ++(0,-0.35) -| (mt2.north);
\draw[arr] (pp.south) -- ++(0,-0.35) -| (mt3.north);
\draw[arr] (pp.south) -- ++(0,-0.35) -| (mt4.north);

% ═══════════════════════════════════════════════════════════════
% MERGED TOKEN SEQUENCE (centred between col 1 and col 3)
% ═══════════════════════════════════════════════════════════════
\def\seqcx{5.5}   % horizontal centre of token row

% place 7 tokens: et  ec  m1  m2  ...  mT
\foreach \i/\lbl in {1/$\mathbf{e}_c$, 2/$\mathbf{e}_t$,
                     3/$\mathbf{m}_1$, 4/$\mathbf{m}_2$,
                     5/${\cdots}$,     6/$\mathbf{m}_T$}{
    \node[tokbox] (st\i) at ({2.0+(\i-1)*1.35}, \re) {\footnotesize\lbl};
}

% ── BUS CONNECTION: upper tokens → sequence row ─────────────────
% All upper tokens drop to a shared horizontal bus, then arrows
% fall cleanly from the bus into each sequence token.
\coordinate (busref) at (0, -5.2);  % midway between \rd=-4.1 and \re=-6.3

% verticals down to bus (no arrowhead)
\foreach \nd in {ec, et, mt1, mt2, mt3, mt4}{
    \draw[thick] (\nd.south) -- (\nd.south |- busref);
}
% single horizontal bus line
\draw[thick] (ec.south |- busref) -- (mt4.south |- busref);

% arrows from bus down into sequence tokens
\foreach \i in {1,2,3,4,5,6}{
    \draw[arr] (st\i.north |- busref) -- ([yshift=2pt]st\i.north);
}

% ═══════════════════════════════════════════════════════════════
% TRANSFORMER ENCODER
% ═══════════════════════════════════════════════════════════════
\node[transbox, minimum width=12.5cm, text width=12cm, align=center]
    (trans) at (\seqcx, \rf)
    {\textbf{Transformer Encoder}
     % \quad{\footnotesize 4 layers $\times$ 8 heads,\; $d_{\mathrm{model}}=512$,\; FFN\,=\,2048,\; GELU}
     };

% arrows: sequence → transformer
\foreach \i in {1,2,3,4,5,6}{
    \draw[arr] (st\i.south) -- (st\i.south |- trans.north);
}

% ═══════════════════════════════════════════════════════════════
% OUTPUT
% ═══════════════════════════════════════════════════════════════
\node[outbox, minimum width=4cm] (outmlp) at (\seqcx, \rg)
    {2-layer MLP (per-frame)};
\node[outbox, minimum width=4cm, text width=3.8cm, align=center]
    (x0hat) at (\seqcx, \rh)
    {$\hat{\mathbf{x}}_0 \in \mathbb{R}^{T \times D}$};

\draw[arr] (trans)  -- (outmlp);
\draw[arr] (outmlp) -- (x0hat);

\end{tikzpicture}
\caption{\textbf{Model architecture overview.}
Three branches prepare input tokens in parallel:
(left)~the gloss label---optionally concatenated with phonological attributes from ASL-LEX 2.0---is encoded by a frozen text encoder (CLIP or T5) and projected by a 3-layer MLP to form the condition token $\mathbf{e}_c$;
(center)~the diffusion timestep is embedded via sinusoidal encoding and a 2-layer MLP to form $\mathbf{e}_t$;
(right)~the noisy motion $\mathbf{x}_t \in \mathbb{R}^{T\times D}$ is linearly projected with learned positional encoding to form per-frame tokens $\mathbf{m}_1,\ldots,\mathbf{m}_T$.
The concatenated sequence is processed by a 4-layer Transformer Encoder, and the output motion tokens are decoded by a per-frame MLP to predict the clean motion $\hat{\mathbf{x}}_0$.}
\label{fig:architecture}
\end{figure*}

The condition token $\mathbf{e}_c$ is produced by encoding the conditioning input and projecting it to the model dimension via a 3-layer MLP with GELU activations (see Figure~\ref{fig:architecture}). In this work, we focus on the generation of single glosses (words). We explore six conditioning variants that differ in the choice of text encoder and type of conditioning. 

\paragraph{Gloss-only conditioning.}
The gloss label (\eg, ``BOOK'') is encoded by a frozen pretrained text encoder to obtain a fixed-dimensional embedding, which is then projected to the model dimension to produce $\mathbf{e}_c \in \mathbb{R}^{512}$.

\paragraph{Gloss + attribute conditioning.}
Phonological attributes from ASL-LEX 2.0 are concatenated with the gloss
label as a structured text string and encoded through the same frozen text
encoder and projection. This design keeps the architecture identical across
conditioning modes, isolating the effect of attribute information.

We select 8 attributes as conditioning signals: Handshape, Sign Type,
Path Movement, Major Location, Minor Location, Nondominant Handshape,
Selected Fingers, and Flexion.
Among these, three attributes (Handshape, Nondominant Handshape, and
Selected Fingers) are annotated with symbolic notation in ASL-LEX 2.0.
For example, the raw ASL-LEX symbols for \textsc{book}---``\texttt{open\_b}''
(handshape) and ``\texttt{imrp}'' (selected fingers)---are translated to
``\texttt{all four fingers extended together thumb extended}'' and
``\texttt{index and middle and ring and pinky}'' respectively.

To study the effect of this notation format, we evaluate two variants:
\textit{symbolic}, which uses the raw ASL-LEX symbols directly, and
\textit{mapped}, which uses their natural language translations.
The remaining five attributes are identical across both variants.
The effect of notation format on generation quality is analyzed in
Table~\ref{tab:ablation}. Glosses without an exact match in ASL-LEX 2.0 fall back to gloss-only conditioning.
% {\bf JK is this example exact? YES}

\paragraph{Text encoder variants.}
We investigate two pretrained text encoders: (1)~{CLIP} ViT-B/32~\cite{radford2021learning}, which produces a 512-d pooled embedding via its text encoder trained with visual-language contrastive learning; and (2)~{T5-base}~\cite{raffel2020exploring}, a seq2seq language model whose encoder outputs are mean-pooled over non-padding tokens to produce a 768-d embedding. Both encoders are frozen during training and only the MLP projection is learned.

% ======================================================================
% 4. EXPERIMENTS
% ======================================================================
\section{Experiments}
\label{sec:experiments}
In this work, we focus only on gloss-level generation and explore the use of phonological attributes that are available as annotations associated with glosses in the ASL-LEX 2.0 database. Gloss is a written label associated with a sign and is sometimes equivalent to a single word or more than one word.   

% \subsection{Dataset}

\paragraph{ASL3DWord.}\label{sec:dataset}
We train and evaluate our model on ASL3DWord~\cite{dong2024signavatar}, a word-level ASL dataset derived from WLASL~\cite{li2020word}, using SMPL-X parameters as a representation.  The dataset contains 103 gloss classes with corresponding 3D motion sequences obtained by fitting an SMPL-X model to video.  For direct comparison, we use the same train/test split for learning the diffusion model as SignAvatar.

\paragraph{ASL-LEX 2.0.}
For phonological conditioning, we use ASL-LEX 2.0~\cite{sehyr2021asl}, a lexical database of 2,723 ASL signs annotated with 22 phonological dimensions. We select 8 attributes as conditioning signals, covering handshape, finger configuration, location, movement, and flexion. The majority of 103 glosses in ASL3DWord have matching entries in ASL-LEX 2.0, providing phonological labels for conditioning.

\subsection{Evaluation Metrics}
\label{sec:metrics}

We evaluate generation quality using both model-free and model-based metrics, following standard protocols in motion generation~\cite{guo2020action2motion,tevet2023human}.

\paragraph{Model-based metrics.}
We train a motion classifier on the training set and use it to compute: (1)~\textbf{FID}: Fr\'{e}chet Inception Distance in the classifier's feature space, measuring distributional similarity between generated and ground-truth motions (lower is better); (2)~\textbf{Accuracy}: classification accuracy of generated motions by the pretrained classifier, measuring sign recognition performance (higher is better).

\paragraph{Model-free metrics.}
(1)~\textbf{KNN Accuracy}: $k$-nearest-neighbor accuracy in feature space, measuring whether generated samples are close to ground-truth samples of the same class (higher is better); (2)~\textbf{Diversity}: average pairwise distance among all generated samples, should be close to ground truth; (3)~\textbf{Multimodality}: average pairwise distance among samples of the same class, should be close to ground truth.

\paragraph{Variance Ratio.}
We introduce \textbf{Variance Ratio} as a body-part-specific metric: the ratio of per-joint variance in generated motions to that in ground-truth motions, computed separately for arms, left hand, right hand, and torso. A ratio of 1.0 indicates that the generated motions match ground-truth articulation amplitude; values below 1.0 indicate under-articulation, and values above 1.0 indicate over-articulation. This metric is particularly relevant for sign language, where hand and arm dynamics carry primary linguistic information.

\subsection{Implementation Details}
\label{sec:impl}

The Transformer Encoder denoiser has 4 layers, 8 heads, and model dimension 512. We train with AdamW~\cite{loshchilov2019decoupled} (learning rate $10^{-4}$, weight decay $10^{-2}$) using cosine annealing, batch size 100, mixed precision (fp16), and gradient clipping at 1.0. Training runs for 500 epochs on a single GPU via the HuggingFace Accelerate library. The diffusion process uses 1000 time steps with a cosine noise schedule; inference uses DDIM with 50 steps. The baseline method SignAvatar uses the official implementation trained for 5000 epochs on the same dataset.

\subsection{Results}
\label{sec:main_results}

\paragraph{Diffusion Baseline vs.\ SignAvatar.}
\label{sec:baseline_results}

\begin{figure*}[t]
\centering
\setlength{\tabcolsep}{2pt}
\begin{tabular}{r @{\hspace{4pt}} l}
  \rotatebox[origin=c]{90}{\footnotesize\textbf{GT}} &
  \includegraphics[width=0.95\textwidth]{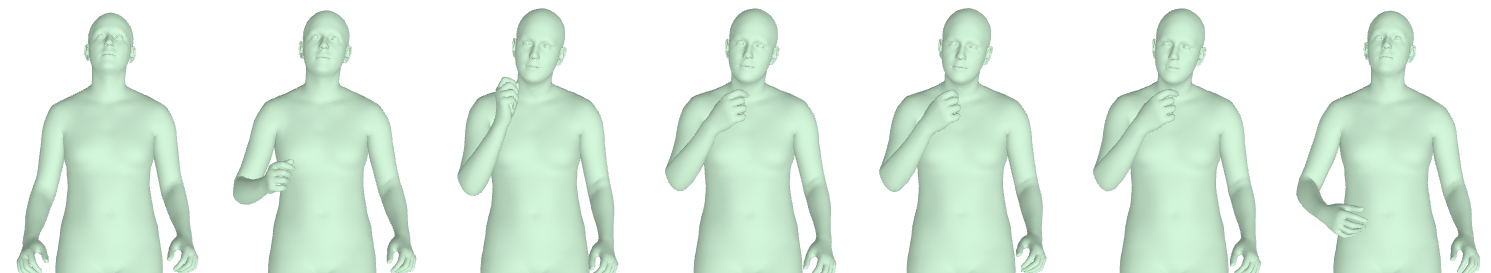} \\[2pt]
  \rotatebox[origin=c]{90}{\footnotesize\textbf{SignAvatar}} &
  \includegraphics[width=0.95\textwidth]{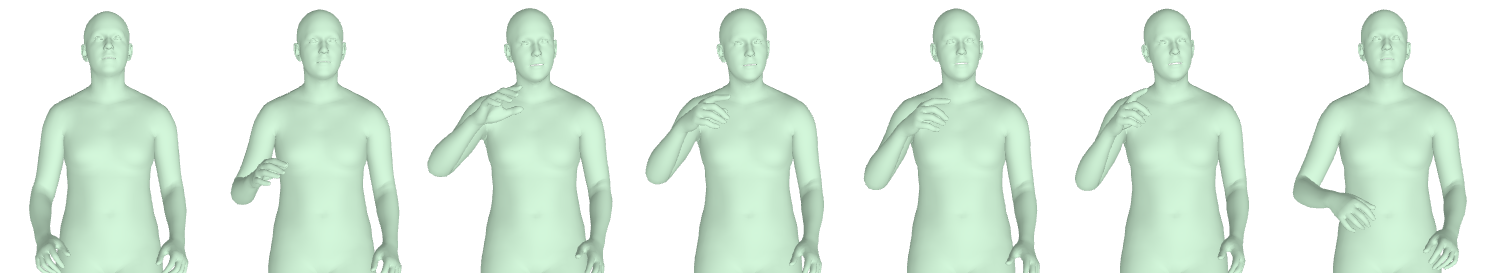} \\[2pt]
  \rotatebox[origin=c]{90}{\footnotesize\textbf{Ours}} &
  \includegraphics[width=0.95\textwidth]{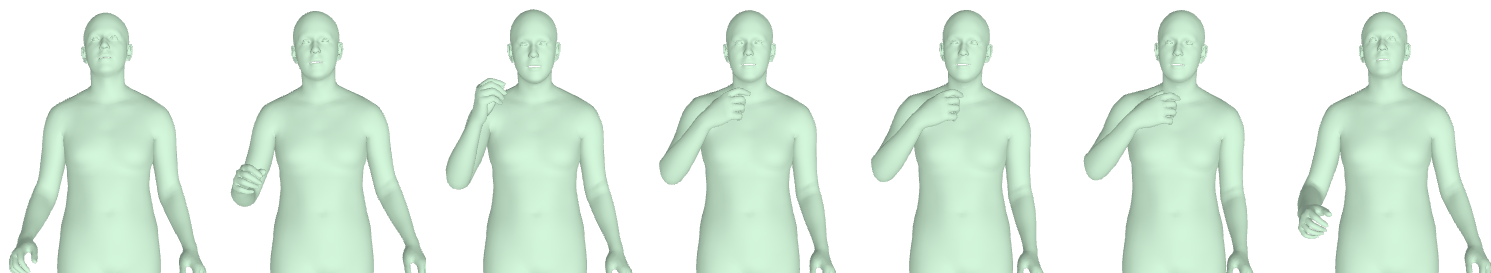} \\
\end{tabular}
\caption{\textbf{Qualitative comparison} for the gloss \textit{cool}. We show evenly-sampled keyframes.
  \textit{Top row}: ground-truth motion from ASL3DWord. \textit{Middle row}: SignAvatar~\cite{dong2024signavatar} generation. \textit{Bottom row}: our CLIP gloss-only diffusion model.}
\label{fig:qualitative}
\end{figure*}

Table~\ref{tab:main_results} compares our diffusion baseline (gloss-only, CLIP encoder) against SignAvatar~\cite{dong2024signavatar} that trains a conditional VAE with the same conditioning. Our model achieves substantially higher gloss discriminability: model-based accuracy of 0.838 vs.\ 0.758, and KNN accuracy of 0.378 vs.\ 0.271. Model-based FID is comparable (62.44 vs.\ 62.23). Diversity closely matches ground truth (27.44 vs.\ GT 27.57), indicating that the diffusion model generates well-calibrated output without mode collapse, while SignAvatar shows somewhat reduced diversity (25.63). Qualitative examples are shown in Figure~\ref{fig:qualitative}.

\begin{table}[t]
\centering
\caption{\textbf{Comparison with SignAvatar.} Our diffusion baseline outperforms the CVAE-based SignAvatar on discriminability metrics while achieving comparable FID. GT diversity and multimodality shown for reference.}
\label{tab:main_results}
\smallskip
\resizebox{\columnwidth}{!}{%
\begin{tabular}{l c c c c c}
\toprule
\textbf{Method} & \textbf{FID}$\downarrow$ & \textbf{Acc.}$\uparrow$ & \textbf{KNN}$\uparrow$ & \textbf{Div.} & \textbf{MM} \\
\midrule
GT & 0.0 & 0.755 & 0.549 & 27.57 & 11.68 \\
\midrule
SignAvatar~\cite{dong2024signavatar} & 62.23 & 0.758 & 0.271 & 25.63 & 10.06 \\
\textbf{Ours (Diff.)} & 62.44 & \textbf{0.838} & \textbf{0.378} & \textbf{27.44} & 9.33 \\
\bottomrule
\end{tabular}
}
\end{table}

\paragraph{Text Encoder $\times$ Conditioning $\times$ Notation Format Ablation.}
\label{sec:ablation}

Table~\ref{tab:ablation} presents the full ablation over text encoders (CLIP vs.\ T5), conditioning modes (gloss-only vs.\ gloss+attributes), and attribute notation format (symbolic vs.\ mapped). This ablation isolates the effects of each factor while keeping the diffusion architecture, training procedure, and loss function identical.
\begin{table}[t]
\centering
\caption{\textbf{Text encoder $\times$ conditioning ablation.} CLIP excels at gloss-only encoding but collapses with symbolic attribute notation; after mapping to natural language, CLIP+Attr achieves the best FID overall. T5 is largely unaffected by the notation format. Div.\ and MM are best when closest to GT (27.57 and 11.68 respectively).}
\label{tab:ablation}
\smallskip
\resizebox{\columnwidth}{!}{%
\begin{tabular}{l l l c c c c c}
\toprule
\textbf{Encoder} & \textbf{Cond.} & \textbf{Attr. Format} & \textbf{FID}$\downarrow$ & \textbf{Acc.}$\uparrow$ & \textbf{KNN}$\uparrow$ & \textbf{Div.} & \textbf{MM} \\
\midrule
CLIP & Gloss      & --       & 62.44          & \textbf{0.838} & \textbf{0.378} & 27.44          & 9.33  \\
CLIP & Gloss+Attr & Symbolic & 127.95         & 0.201          & 0.097          & 26.61          & 12.67 \\
CLIP & Gloss+Attr & Mapped   & \textbf{55.11} & 0.794          & 0.345          & \textbf{27.51} & 10.44 \\
\midrule
T5   & Gloss      & --       & 66.70          & 0.496          & 0.304          & 27.22          & 12.18 \\
T5   & Gloss+Attr & Symbolic & 63.66          & 0.525          & 0.310          & 27.31          & 12.23 \\
T5   & Gloss+Attr & Mapped   & 63.58          & 0.546          & 0.310          & 27.30          & \textbf{11.75} \\
\bottomrule
\end{tabular}
}
\end{table}
CLIP benefits dramatically from natural language mapping (accuracy 
0.201 $\rightarrow$ 0.794, FID 127.95 $\rightarrow$ 55.11), while 
T5 remains largely unaffected (accuracy 0.525 $\rightarrow$ 0.546). 
We analyze these contrasting behaviors in Section~\ref{sec:analysis}.

\paragraph{Cross-Method Comparison.}
\label{sec:cross_comparison}
Combining results from Table~\ref{tab:main_results} and Table~\ref{tab:ablation}, CLIP with mapped phonological attributes (CLIP+Attr Mapped) achieves the best FID overall (55.11) and outperforms SignAvatar across all reported metrics, demonstrating that effective attribute conditioning further improves over the gloss-only diffusion baseline.

\paragraph{Variance Ratio.}
\label{sec:variance}

Table~\ref{tab:variance} reports variance ratios by body part for mapped
attribute variants; detailed analysis is provided in Section~\ref{sec:analysis}.

\begin{table}[t]
\centering
\caption{\textbf{Variance Ratio} (generated/GT variance per body part; 
ideal = 1.0). SignAvatar consistently under-articulates; all diffusion 
models show mild over-articulation in hands and arms.}
\label{tab:variance}
\smallskip
\resizebox{\columnwidth}{!}{%
\begin{tabular}{l c c c c c}
\toprule
\textbf{Body Part} & \textbf{SAvatar} & \textbf{CLIP+G} & \textbf{CLIP+GA} & \textbf{T5+G} & \textbf{T5+GA} \\
\midrule
Arms   & 0.620 & 1.289 & \textbf{1.278} & 1.337 & 1.370 \\
L-Hand & 0.495 & 1.470 & 1.437          & \textbf{1.401} & 1.449 \\
R-Hand & 0.622 & 1.348 & 1.352          & \textbf{1.262} & 1.347 \\
Torso  & 0.675 & 0.756 & 0.747          & 0.774 & \textbf{0.831} \\
\bottomrule
\end{tabular}
}
\smallskip
{\footnotesize SAvatar = SignAvatar, G = Gloss-only, GA = Gloss+Attributes (mapped). Bold = closest to 1.0.}
\end{table}

% ======================================================================
% 5. ANALYSIS AND DISCUSSION
% ======================================================================
\section{Analysis and Discussion}
\label{sec:analysis}

The ablation in Table~\ref{tab:ablation} reveals a striking asymmetry
between CLIP and T5 in their response to attribute notation format.
In this section, we discuss the underlying reasons and broader implications.

\paragraph{The Role of Input Representation for CLIP.}
\label{sec:clip_failure}
CLIP's text encoder is trained via contrastive learning on a large dataset of web image-cation pairs aligning text
with visual semantics. 
It is effective for encoding short common
English words as is the case for single-word glosses in our vocabulary.
However, when symbolic ASL-LEX symbols are appended as attributes
(e.g., \texttt{open\_b}, \texttt{imrp}), performance collapses
dramatically: accuracy drops from 0.838 to 0.201 and FID doubles from
62.44 to 127.95. After translating these symbols to natural language,
CLIP+Attr recovers reaching accuracy of 0.794 and FID improves
to 55.11, surpassing all other variants. This reveals that the failure
was not a fundamental limitation of CLIP on structured attributes, but
a consequence of out-of-distribution symbolic tokens. Furthermore,
semantically unrelated glosses can share identical attribute profiles---
for instance, \textsc{book} and \textsc{walk} are identical across all
8 conditioning attributes---so the appended attribute strings dominate
the pooled embedding and overwhelm the gloss token, making conditions
for different signs nearly indistinguishable.
% {\bf JK did you actually check this ? YES. I replaced 'HOUSE' with 'WALK', and these two ('book' and 'walk') share all the attributes.}

\paragraph{Why T5 Is Unaffected by the Mapping.}
\label{sec:t5_recovery}
T5, trained on diverse text-to-text tasks, handles structured strings
naturally via its sub-word tokenizer, which treats symbolic symbols such
as \texttt{open\_b} and natural language equivalents as equally distinct
token sequences. This explains why mapping has almost no effect on T5+Attr
performance (accuracy 0.525 $\rightarrow$ 0.546, FID 63.66 $\rightarrow$
63.58). Critically, T5 gloss-only (0.496) performs substantially worse
than CLIP gloss-only (0.838), as single common English words under-utilize
T5's seq2seq capacity. The asymmetric response to mapping---dramatic for
CLIP, negligible for T5---confirms that the two encoders differ
fundamentally in how they process symbolic versus natural language input,
and that the text encoder, not the diffusion architecture, is the primary
bottleneck for attribute integration.

\paragraph{Variance Ratio Analysis.}
\label{sec:variance_paradox}
Table~\ref{tab:variance} reports variance ratios by body part for 
mapped attribute variants. SignAvatar consistently under-articulates 
(ratios 0.495--0.675), particularly for hands. All four diffusion 
variants show mild over-articulation in hands and arms (1.262--1.470), 
with no notable outlier among them. The over-articulation pattern is 
consistent across conditioning modes, suggesting it reflects a 
property of the diffusion model rather than the conditioning signal. 
Improving articulation precision remains an open challenge, and may 
benefit from more fine-grained loss supervision at the joint level 
or stronger phonological conditioning via structured tensor pathways.

% \subsection{Potential Factors in Conditioning Failure}
% \label{sec:root_causes}
\noindent
The performance is further affected by additional factors that 
we briefly discuss below.

\paragraph{Noisy attribute--video alignment.}
ASL-LEX 2.0 provides phonological attributes for the \textit{canonical citation form} of each sign. However, WLASL~\cite{li2020word} videos contain multiple signing variants per gloss, where different signers may use different hand shapes, locations, or movement patterns for the same sign. Only a subset of WLASL videos match the ASL-LEX annotations. This creates a label noise problem: the model receives identical attribute conditioning for videos with genuinely different phonological realizations, which directly undermines the conditioning signal.

% \paragraph{Shared attributes collapse gloss identity.}
% Many glosses share subsets of phonological attributes (\eg, similar handshapes or locations). When attributes are encoded as a concatenated text string, different glosses with overlapping attribute profiles may produce near-identical condition embeddings, causing the model to generate similar motions for distinct signs. This effect is likely amplified by pooling operations in text encoders, which compress the full attribute string into a single vector without guaranteeing that each dimension is independently preserved.
% {\bf JK - maybe we can omit this as this paragraph is captured to some extent by comments about text encoder}

\paragraph{Toward Structured Tensor Conditioning.}
\label{sec:future}

While natural language mapping substantially improves CLIP-based attribute conditioning, our findings further motivate replacing the text-string encoding pipeline with a structured tensor conditioning approach, where the gloss word and each phonological attribute are encoded through independent pathways---\eg, a learnable gloss embedding lookup combined with per-attribute embeddings injected via FiLM~\cite{perez2018film} modulation.
This design avoids dependence on text encoder properties, preserves the compositional independence of each phonological dimension, and provides sufficient representational capacity for both sign identity and attribute structure simultaneously.

% ======================================================================
% 6. CONCLUSION
% ======================================================================
\section{Conclusion}
\label{sec:conclusion}

We present a diffusion-based framework for sign language motion generation and conduct a systematic investigation of phonological attribute conditioning. Our MDM-style diffusion baseline, operating on SMPL-X upper-body motion in 6D rotation representation, outperforms SignAvatar on gloss discriminability while achieving comparable distributional quality; furthermore, CLIP with mapped phonological attributes outperforms SignAvatar across all metrics, demonstrating that effective attribute conditioning yields additional gains beyond the gloss-only baseline.
Through an ablation over text encoders, conditioning modes, and attribute 
notation format, we show that translating symbolic ASL-LEX notations 
to natural language is a necessary condition for effective CLIP-based 
attribute conditioning, and that CLIP and T5 respond asymmetrically to 
this translation.

We further show that input representation is the primary bottleneck for 
text-encoder-based attribute conditioning: CLIP requires semantic 
alignment between input text and its training distribution, while T5's 
token-level structure makes it robust to notation format. The persistent 
noisy alignment between WLASL video variants and ASL-LEX canonical 
annotations remains a limiting factor regardless of notation format. 
These findings motivate structured tensor conditioning, where gloss 
identity and phonological attributes are encoded through independent, 
dimension-preserving pathways---a direction we believe represents a 
promising path toward controllable, linguistically grounded sign 
language generation.

\paragraph{Limitations.}
Our study is conducted on a 103-class word-level vocabulary; scaling to larger vocabularies and sentence-level generation remains future work. The noisy alignment between WLASL and ASL-LEX 2.0 limits the conclusions that can be drawn about attribute conditioning; a curated dataset with verified phonological labels would enable stronger evaluation. Structured tensor conditioning remains to be implemented and empirically validated as an immediate next step.

\paragraph{Broader Impact.}
This work aims to advance technology for deaf or  hard-of-gearing communication. We acknowledge that sign language generation systems should be developed in close collaboration with the deaf and hard-of-hearing community to ensure cultural sensitivity and linguistic accuracy.

% ======================================================================
% REFERENCES
% ======================================================================
{
    \small
    \bibliographystyle{ieeenat_fullname}
    \bibliography{main}
}

\end{document}